Computer Modeling in
Engineering & Sciences

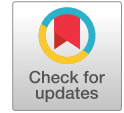

Tech Science Press

**ARTICLE**

# Methods for the Segmentation of Reticular Structures Using 3D LiDAR Data: A Comparative Evaluation

**Francisco J. Soler Mora**[1,*] , **Adrián Peidró Vidal**[1] , **Marc Fabregat-Jaén**[1] , **Luis Payá Castelló**[1,2] **and Óscar Reinoso García** [1,2]

[1]Engineering Research Institute of Elche (I3E), Avenida de la Universidad, s/n, Elche, 03202, Spain
[2]Valencian Graduate School and Research Network of Artificial Intelligence (ValgrAI), Camí de Vera S/N, Edificio 3Q, Valencia, 46022, Spain
*Corresponding Author: Francisco J. Soler Mora. Email: f.soler@umh.es

**ABSTRACT:** Reticular structures are the basis of major infrastructure projects, including bridges, electrical pylons and airports. However, inspecting and maintaining these structures is both expensive and hazardous, traditionally requiring human involvement. While some research has been conducted in this field of study, most efforts focus on faults identification through images or the design of robotic platforms, often neglecting the autonomous navigation of robots through the structure. This study addresses this limitation by proposing methods to detect navigable surfaces in truss structures, thereby enhancing the autonomous capabilities of climbing robots to navigate through these environments. The paper proposes multiple approaches for the binary segmentation between navigable surfaces and background from 3D point clouds captured from metallic trusses. Approaches can be classified into two paradigms: analytical algorithms and deep learning methods. Within the analytical approach, an ad hoc algorithm is developed for segmenting the structures, leveraging different techniques to evaluate the eigendecomposition of planar patches within the point cloud. In parallel, widely used and advanced deep learning models, including PointNet, PointNet++, MinkUNet34C, and PointTransformerV3, are trained and evaluated for the same task. A comparative analysis of these paradigms reveals some key insights. The analytical algorithm demonstrates easier parameter adjustment and comparable performance to that of the deep learning models, despite the latter's higher computational demands. Nevertheless, the deep learning models stand out in segmentation accuracy, with PointTransformerV3 achieving impressive results, such as a *Mean Intersection Over Union* (mIoU) of approximately 97%. This study highlights the potential of analytical and deep learning approaches to improve the autonomous navigation of climbing robots in complex truss structures. The findings underscore the trade-offs between computational efficiency and segmentation performance, offering valuable insights for future research and practical applications in autonomous infrastructure maintenance and inspection.

## 1 Introduction

The durability and mechanical resilience of lattice structures have established them as indispensable components in a multitude of contemporary infrastructure systems. From telecom towers and power line pylons to scaffolding and bridge frames (Fig. 1), lattice designs offer a high strength-to-weight ratio, making them ideal for large-scale constructions that require both durability and reliability.

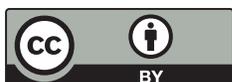

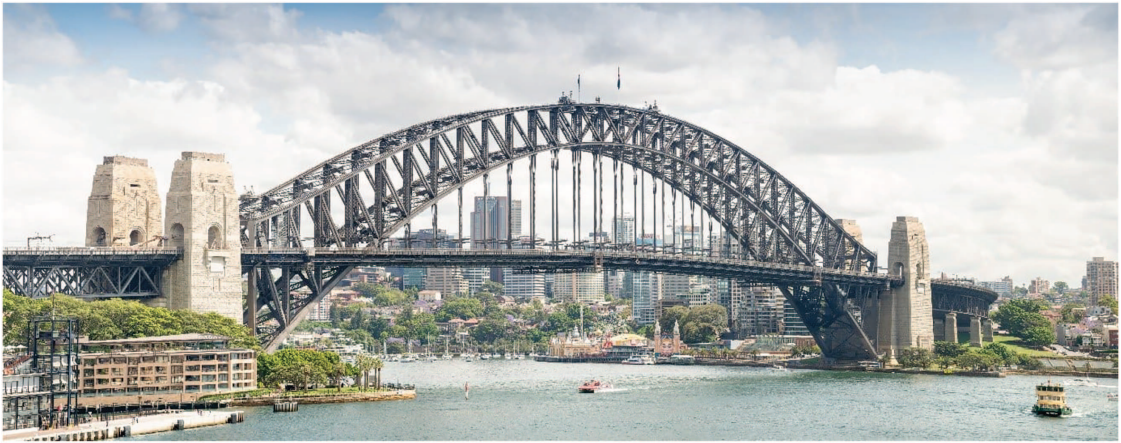

**Figure 1:** Sydney harbour bridge

The geometric arrangement of multiple interconnected elements of these structures allows an efficient distribution of loads and forces across multiple nodes, reducing stress concentration and enhancing overall stability. This combination of characteristics has led to their widespread adoption in engineering and architectural designs, where their inherent stability and adaptability contribute to the resilience and longevity of critical infrastructure.

Their metallic composition and common placement in outdoor areas require regular maintenance and inspection to ensure their integrity. These tasks, traditionally carried out by human operators, expose them to significant risks, including falls, high voltages, and other hazards.

One of the principal methods used to mitigate the risk to operators is the use of robots, either for the complete performance of the task or to provide assistance. During recent decades, a variety of solutions have been proposed to increase the automation of inspection and maintenance tasks of complex truss structures such as climbing robots [1,2] and drones [3–5].

Climbing robots generally have a longer lifespan, payload, stability, and are less affected by environmental factors [6,7]. In particular, bipedal climbing robots provide considerable flexibility, agility, and payload capacity, allowing them to traverse the interior and exterior of lattice structures even when these structures have a high density of lattice elements. This capability arises from their numerous degrees of freedom, precise control mechanisms, and ability to easily navigate around obstacles.

In contrast, drones are faster, easier to deploy (it is not necessary for the operator to be in close proximity to the structure in order to initiate the task), and are not dependent on the surface of the structure. However, this type of device is generally prone to difficulties when navigating within high-density structures, suffers instability in confined spaces due to proximity to structural elements, there is an increased risk of crashing, colliding, or falling debris, and has a limited battery [6,7].

The complexity of reticular structures poses unique challenges when it comes to modelling, analysing, and inspecting them, particularly in 3D environments, mainly due to their complex geometry, overlapping elements and potential occlusions. Consequently, accurate segmentation of spatial information is essential for autonomous climbing robots to perform inspection, maintenance, and navigation tasks successfully.

The acquisition of spatial information related to the environment requires sensors capable of measuring depth. With the emergence of more affordable LiDAR sensors [8] and RGB-D cameras [9], spatial information has been introduced to address 3D segmentation.

RGB-D cameras are sensors that capture both colour (RGB) and depth (D) information from a scene. This is typically achieved by stereo vision in combination with a time-of-flight (TOF) sensor. The combination of stereo vision and the TOF sensor allows building robust and accurate depth maps [10]. The TOF sensor enables the robot to construct a 3D depth map even when the RGB information of the stereo vision is not enough for the 3D reconstruction. Despite this, RGB-D cameras are susceptible to fluctuations due to the lighting conditions and have a relatively narrow operational range, typically between 0.5 and 5 m.

LiDAR, which stands for Light Detection and Ranging, uses laser pulses to measure distances to objects or surfaces. Its operational principle is based on the emission of a series of rapid laser pulses, and timing how long it takes for each pulse to bounce back after interacting with an object. LiDAR systems create detailed 3D maps or "point clouds" of the environment with high accuracy. The ability to capture high-resolution, long-range (0.5–100 m), and accurate spatial information under various lighting and weather conditions makes them more effective and advantageous sensors in mobile robot applications [11].

However, the large amount of information provided by these sensors can be both an asset and a liability. Large volumes of data, which often include an important amount of noise and irrelevant information, can negatively impact the performance of navigation algorithms. To address this, various filters are typically employed to reduce and refine the sensor data, such as voxel filtering for down-sampling and statistical methods for outlier removal. After filtering, segmentation plays a crucial role, as it allows identifying different elements within the environment, enabling autonomous systems to make informed decisions for navigation and task execution [12].

In this work, various methods are presented to achieve a binary segmentation between truss structures and background, which will facilitate subsequent works on modelling, mapping, and navigating through these types of structures. Neural networks have been widely used for 2D segmentation tasks in recent years [13–15].

This paper is structured as follows. First, an overview of the related work is presented in Section 2. After this, Section 3 describes the robotic platform. Section 4 outlines the neural network models studied for the segmentation task. Section 5 presents an analytical algorithm to perform a similar segmentation using conventional methods. Section 6 describes the experiments and discusses their results. The study concludes with Section 7, which presents a brief overview of the key findings and proposes future research directions.

The repository with the source code is publicy available at https://github.com/Urwik/rs_seg_methods.git (accessed on 20 May 2025).

## 2 Related Work

Although numerous studies have been conducted in the realm of structural inspection, autonomous navigation through these environments remains poorly studied. Due to their intrincated nature, the majority of the work related to climbing robots for truss structures has focused on the mechanical design and control [9,16,17]. Furthermore, a significant portion of the work in this field focuses on the visual analysis of structural defects or structural health monitoring (SHM) [18–21]. Consequently, most of the work in this field requires human intervention to guide the robot across the structure until the target points.

In addition to the inherent limitations, there are also few examples of research works using autonomous robots on generic reticular structures. The most comparable investigations are those focused on the inspection of metallic bridges. Table 1 summarizes the most relevant works similar to the field of segmentation of reticular structures.

### 2.1 Inspection and Navigation

Two principal methods for the inspection or maintenance of this type of structure are drones [22,23] and climbing robots [24,25]. Drones facilitate faster and easier exploration, though their effectiveness is often limited by the difficulty of navigating among the bars that compose the structures [6]. In contrast, climbing robots are designed to adhere to the structure, offering greater ease, precision, and safety when manoeuvring through complex and narrow environments such as lattice structures [7].

As stated previously, most current research is focused on the development of the robot itself or on the detection of faults, rather than the creation of a model of the structure that enables the automation of the task [26–28]. In the existing literature, there are only a few works that aim to provide a structure model to automate the inspection or maintenance of a structure [29,30]. Lin et al. [29] propose a method for the detection of trusses using a sliding window approach. However, the construction of a complete map of the environment before the application of the algorithm limits the scalability of the method to large structures. Bui et al. [30] propose a novel control framework to minimize human involvement by a switching control that allows the robot to alternate its configurations according to the environment. The control system uses 3D point clouds to identify the nearest flat surface to which the robot can attach. Nevertheless, this method is only capable of detecting the closest planar surface, regardless of the surrounding environment or multiple available surfaces.

### 2.2 Dataset Generation

In order to address the task of segmenting the environment and identifying faults, there is a growing trend toward the use of emerging techniques such as neural networks. These methods require a substantial amount of data for their implementation.

The field of study addressed in this work is still in its early stages and, as a result, there is a dearth of extensive databases that could facilitate the advancement of this research line. Consequently, some researchers have developed novel databases, often resorting to synthetic environments due to the considerable cost associated with capturing and labelling real data. This cost increases when the objective environments are typically found in large infrastructures at high altitudes and in hazardous locations. Therefore, acquiring these data is a costly and risky undertaking. Failure in navigation or adhesion of the robot can cause a fall of the robot, posing serious risks to the robot and the operators.

Some works [31–34] employ different methodologies to generate synthetic datasets that can be used to develop algorithms for the inspection and segmentation of structural elements. One of the advantages of this type of database is that it facilitates the automation of data labelling, thereby enabling the direct acquisition of ground truth from the simulation environment. The following paragraphs provide a description of the aforementioned works.

The study by Cheng et al. [31] utilizes Blender in conjunction with a Python API to generate a database of life-like images set within environments featuring six different types of parametrically generated bridges. Following this, the effectiveness of the dataset is verified by training the DeepLabV3Plus neural network [35] for semantic image segmentation, yielding promising results.

In parallel with previous research, Lamas et al. [32] have developed a MATLAB software to create different types of reticular bridges and to implement diverse profiles of the bridge structure. The software is tailored to train the JSNet network model [36] to achieve semantic segmentation and segmentation by instance of each constituent element within the point cloud.

The approach outlined in [33] employs a combination of simulated data and real 3D LiDAR data to increase the number of point cloud examples required for the training of deep learning methods. The authors introduce two methodologies for data augmentation, illustrating that the integration of such processes with the initial database yields improved results across all aspects of segmentation.

In addition, Jing et al. [34] address the dearth of databases containing structural elements with information generated synthetically. The authors evaluate the performance of their approach using neural networks for semantic segmentation. Their dataset comprises both real and synthetic clouds generated from 3D models. Furthermore, they propose a neural network model (BridgeNet) and compare its performance with other segmentation models, including PointNet++ [37] and RandLA-Net [38].

In the present work a Gazebo plugin has been developed to generate the necessary datasets that allow the implementation and test of the methods for the segmentation of reticular structures. This plugin facilitates the parameterization and automation of the process of generation and labelling of synthetic datasets. Furthermore, the data generated through this method closely resembles the behaviour of the sensor itself, eliminating the need for post-processing to obtain occlusions and noise.

### 2.3 Deep Learning Structure Segmentation

Neural networks stand out as one of the predominant methodologies used for semantic segmentation of objects. Their widespread adoption is attributed to their commendable performance, versatility, and ability to extract distinctive features for various objects. Leveraging this feature extraction capability, they can effectively assign a semantic label to each point within a sensor scan.

Within the domain of structure inspection and maintenance, several studies use neural networks to segment data acquired from LiDAR sensors. This segmentation is commonly used to identify specific elements of the infrastructure for evaluation and to focus attention on these elements.

Ji et al. [39] developed a neural network called *Dual Attention-based Point Cloud Network* (DAPCNet) for multi-class segmentation in tunnel environments using point cloud data. The DAPCNet uses an encoder-decoder architecture with a dual attention module to improve the segmentation. They also propose a loss function called *Facal Cross-Entropy* to address an imbalanced data distribution. They achieve better results than other networks used for segmentation like PointNet [40] and *Dynamic Graph CNN* (DGCNN) [41].

Grandio et al. [42] use PointNet++ [37] to segment four different scenarios with railway data. They use the euclidean coordinates and the intensity value of each point to train the segmentation for eight different semantic classes.

In [43], three network models are evaluated for bridge segmentation. The evaluated models are DGCNN, PointNet, and PointCNN. The best mean intersection over union (mIoU) is achieved by DGCNN (86.85%) closely followed by PointNet (84.29%).

Kim et al. [44] present a PointNet-based neural network for the segmentation in three different classes of bridges. Background is considered a class so no pre-processing step of the cloud is needed.

The present work applies semantic segmentation neural networks to achieve a binary segmentation of metallic structures. To obtain models which are able to generalize, the present work proposes to train the neural networks with simple geometric elements (parallelepipeds) and subsequently evaluate them with complex structures. The method employs single, sparse, and uncorrelated scans. A study has also been conducted on the input features for the networks and how they affect the segmentation efficacy.

### 2.4 Analytical Structure Segmentation

Analytical techniques can be used to achieve segmentation outcomes comparable to those obtained with neural networks. However, these methods often have a limited ability to generalize. Notwithstanding that, their use can provide advantages in some scenarios where the use of dedicated hardware or neural network training is unfeasible due to limited data availability.

In [45], an analytical method for the segmentation of metallic truss bridges is proposed. The data is acquired using a Terrestrial Laser Scanner (TLS). A preprocessing step is executed to homogenize the density of each scan and remove noisy isolated points, improving the next steps of the segmentation. Segmentation is achieved by analysing each face of the bridge (vertical, horizontal, interior) separately. For each direction, a combination of *Principal Component Analysis* (PCA) with *GPU-Density Based Spatial Clustering* GDBSCAN [46] is used to complete the bar segmentation.

Riveiro et al. [47] present a method to segment a dense point cloud map of masonry arch bridges. In this work PCA is used for normal estimation. They first classify points between vertical and non-vertical walls using normal estimation. By transforming the normal vectors into spherical coordinates, a final segmentation is obtained that classifies the vertical walls into different parts from the azimuth histogram.

In their study, Yan et al. [48] use a preprocessing step to remove irrelevant data, extract local point features and downsample the cloud. They propose a segmentation method that sequentially identifies points in different classes. Initially, the scan is classified as either above the bridge or above the ground. Subsequently, depending on the scan type, a distinct combination of PCA, *Random Sample Consensus* (RANSAC) and Euclidean clustering is employed to achieve the final segmentation.

Lu et al. [49] present a method to segment structural components in reinforced concrete bridges. In their approach, a slicing algorithm is applied to separate deck points from pier points. Secondly, surface normals, density histograms and oriented bounding boxes are used to segment pier parts. Finally, a merging stage is applied to reduce over-segmentation results of the bridge.

The present work proposes an analytical method for the binary segmentation of reticular structures. The problem is approached through a two-step algorithm that employs the combination of normal estimation, RANSAC plane detection, and eigenvectors computation to achieve the final segmentation. Furthermore, this work compares the efficiency and effectiveness of the analytical method and the neural network models.

**Table 1:** Summary of the most relevant works similar to the field of segmentation of reticular structures

| Reference | Paradigm | Target | Method | Limitations |
|---|---|---|---|---|
| Ji et al. [39] | Deep learning | Tunnels | DAPCNet | Manually labelled data |
| Grandio et al. [42] | Deep learning | Railways | PointNet++ | Limited labelled test data |
| Kim et al. [43] | Deep learning | Concrete bridges | DGCNN, PointNet, PointCNN | Manually labelled data |
| Kim et al. [44] | Deep learning | Concrete bridges | PointNet-based | Manually labelled data |
| Lamas et al. [45] | Analytical | Truss structures | PCA, DBSCAN | Requires a map |
| Riveiro et al. [47] | Analytical | Masonry arch bridges | Normals, PCA, Azimuth Histogram | Requires a map |

(Continued)

**Table 1 (continued)**

| Reference | Paradigm | Target | Method | Limitations |
|---|---|---|---|---|
| Yan et al. [48] | Analytical | Steel girder bridges | Normals, PCA, RANSAC, K-medoid | Requires a map |
| Lu et al. [49] | Analytical | Concrete bridges | Normals, density histogram, sliding window | Requires a map |

## 3 Robotic Platform

### 3.1 Robot

The methods proposed in this paper are designed for the robot developed in [50]. This robot, designated HyReCRo, is a biped robot developed to perform maintenance and inspection tasks on three-dimensional metallic lattice structures. Each of the robot's legs is composed of two parallel mechanisms with two degrees of freedom (DOF) placed opposite each other. Both parallel modules share a central link, which serves as a slider. The configuration of two parallel modules per leg ensures symmetry and enhances the robot's robustness and payload capacity.

Both legs are connected in series to a common element, designated as the hip, through rotary joints. The connection of parallel modules in series results in the formation of a series-parallel configuration, which will henceforth be referred to as a hybrid configuration. The actual state of the robot prototype can be observed in Fig. 2a. Due to its hybrid architecture, the robot has 10 degrees of freedom (DOF), making it a redundant robot with considerable flexibility to navigate in complex environments like metallic lattice structures.

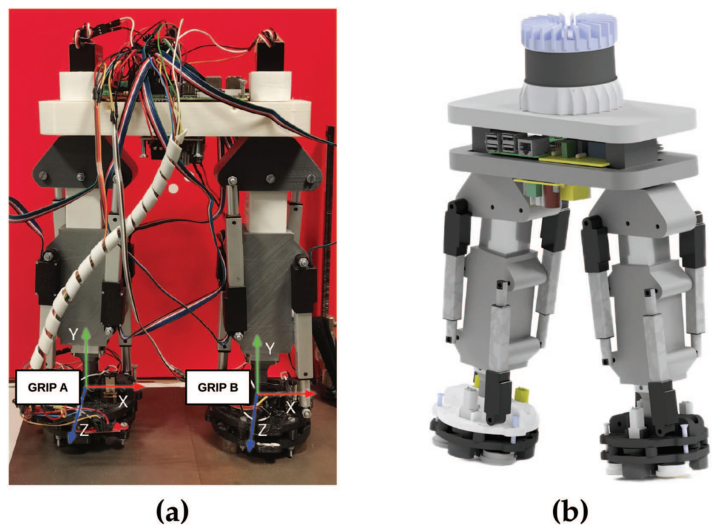

(a)          (b)

**Figure 2:** Example figures of the HyReCRo robot: (**a**) Actual state of the HyReCRo prototype; (**b**) CAD representation of the final HyReCRo robot

The adhesion mechanism is achieved through the use of a permanent magnets system, which is capable of being mechanically switched. This enables minimal power consumption. The robot also has an automatic adhesion process [51].

### 3.2 Sensors

To capture the necessary information from the environment, sensors that provide spatial information have been used, which is common in current applications [8,29]. Within the category of spatial information sensors, two types have been considered: RGB-D cameras and LiDAR sensors. Due to the enhanced quantity of information and accuracy afforded by a LiDAR sensor, this alternative has been selected.

The Ouster OS1 sensor, equipped with 128 channels, serves as a LiDAR sensor of choice. This sensor uses a class 1 laser, enabling its operation without the need for safety equipment. Its primary characteristics encompass a high precision of $\pm 0.03$ m, an extensive operational range spanning from 0.5 to 100 m, and configurable resolutions of 128 × 512, 128 × 1024, and 128 × 2048. Additionally, it offers a configurable frequency of 10/20 Hz, along with a vertical field of view (FOV) ranging from −22.5 to 22.5 degrees and a horizontal FOV of 360 degrees.

In the final prototype of HyReCRo, this sensor would be located at the centre of the hip. In Fig. 2b, a 3D representation of final prototype is shown.

## 4 Deep Learning Segmentation

In this study, some of the most widely used neural network models for 3D semantic segmentation are studied. The aforementioned models include PointNet [40], PoinNet++ [37], MinkUNet34C [52] and PointTransformerV3 [53].

### 4.1 PointNet

PointNet [40] is known for being the first network that enables the use of unorganized or sparse point clouds directly as input. It can be used for a range of tasks, including object classification, object parts segmentation and scene semantic segmentation.

The network is composed of several multi-layer perceptrons (MLPs), which sequentially extract features from the local point to the global scene in a sequential manner. Due to its MLP architecture, the input cloud is restricted to a fixed size, which generally requires preprocessing the initial scan to extract a fixed number of points. Additionally, this network does not consider the neighbourhood of points, as the input is an unordered set of points. Despite the aforementioned features, the system exhibits satisfactory performance in segmentation and classification tasks [54].

The original PointNet architecture is designed to accept a maximum of three features (cartesian coordinates) of input data for each point. However, a relatively straightforward modification of the model has been implemented to enable the number of input features to be parametrized.

### 4.2 PointNet++

PointNet++ [37] appears as an improvement of PointNet. It addresses the shortcoming of only evaluating individual point features but not their neighbourhood. PointNet++ extracts local features, by grouping and applying hierarchically the PointNet backbone to each point neighbourhood. As a center for the local regions, it uses the *Farthest Point Sampling* (FPS) algorithm, which samples the input point cloud to a fixed size while preserving the global geometry of the point cloud.

In order to consider the information of the point neighbourhood, the network groups nearby points and extracts features from them. This process is applied sequentially at different resolutions (obtained by sampling the input cloud). The features extracted from point groups are combined sequentially to include neighbourhood information. Finally, a propagation of the features is applied until the features dimension is equal to the desired number of semantic elements.

The results of the new version prove to be superior, exhibiting greater resilience to input clouds with variable density. However, the training and inference times are significantly longer compared to PointNet, due to the required preliminary sampling and hierarchical learning at different resolutions.

### 4.3  MinkUNet34C

The MinkUNet34C [52] architecture represents a solution to the challenge of reducing the computational time required for 3D convolutions within the Minkowski Engine project [52]. This project introduces a novel approach to convolutions, enabling sparse 3D convolutions that only operate on existing points, in contrast with traditional convolutional network models, which divide the entire space and perform convolutions for all cells regardless of their occupancy.

This model requires two distinct inputs: on the one hand, the coordinates of the points, and on the other hand, a feature vector per point. In order to prevent memory overflow and reduce computational cost, the coordinates of the points are discretized through voxelization. The voxels that contain points are used to build a sparse tensor. The centre of each voxel is used as a convolution centre.

Subsequent operations on the point cloud, such as convolution, are applied to the mean of the features of the points within each voxel. The voxel size must be adjusted according to the specific application, with typical values ranging from 0.05 to 0.15 m. The model employs hierarchical learning, whereby the outputs of the initial blocks serve as inputs for the subsequent blocks. Following an encoder-decoder structure, it assigns a label to each voxel, which is then propagated to all points within that voxel in order to match the size of the original point cloud.

### 4.4  PointTransformerV3

PointTransformerV3 [53] is considered one of the most effective neural network models for semantic segmentation, particularly when applied to well-known datasets such as SemanticKITTI [55] or Scan-Net200 [56]. This model employs an architecture based on self-attention layers (transformers) for the successful completion of tasks such as semantic segmentation, instance segmentation and object detection. In regard to the operational principles of the transformers, the model employs a serialized input, followed by a U-Net framework (encoder-decoder) to apply recursively pooling and self-attention layers. The main contribution of the work is the utilization of space-filling curves (Z-order and Hilbert) for the serialization of the input data. This serialization enhances the efficiency of the model, reducing the computational cost and memory requirements in comparison to the previous work of the autors [57] and other state-of-the-art models, such as MinkUNet34C [52].

## 5  Analytical Method for the Segmentation of Reticular Structures

The proposed method for the analytical segmentation of the structures results in a binary segmentation of lattice structures that emerge from the ground in outdoor environments. The algorithm addresses the challenge of segmenting structures within individual scans, without any correlation between them.

The algorithm employs a two-step approach (Fig. 3). Following the nature of the environment, the first step aims to identify roughly the surface of the ground. The second stage enhances the segmentation by focusing on the contact areas between the structure and the ground.

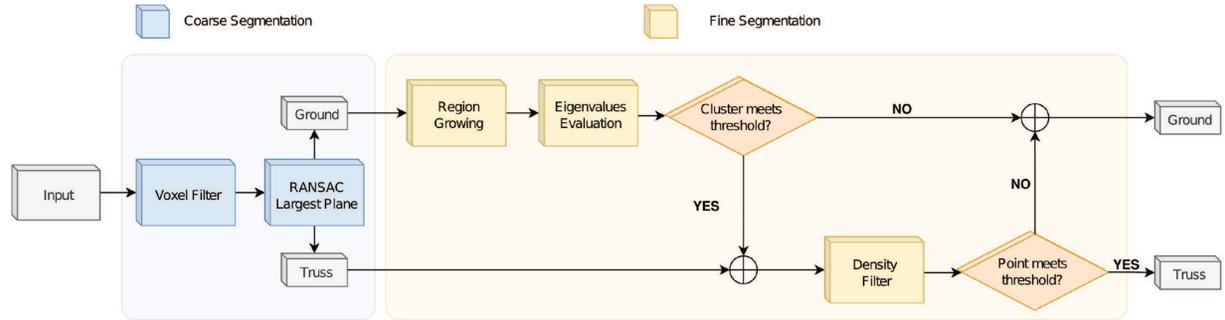

**Figure 3:** Flowchart of the proposed algorithm

The implementation of the point cloud processing was conducted using the Point Cloud Library (PCL) [58].

### 5.1 Coarse Segmentation

The first stage aims to detect a planar model that best fits the ground nearby the structure.

The extraction of a planar model from a point cloud is typically accomplished through the use of the RANSAC algorithm. However, RANSAC is particularly sensitive to the density of points, a factor which requires preprocessing to be applied prior to its use. In this instance, a voxel filter is employed prior to the application of RANSAC, with the objective of extracting the parameters of the plane that optimally align with the ground plane in the vicinity of the truss. Given that the ground is not a perfect planar model, but rather a curved surface, it is necessary to set high RANSAC thresholds in order to achieve the best possible identification of the main ground normal direction. Empirical values of 0.5–1 m are likely to provide sufficient results for most cases.

The voxelized cloud is employed solely for the extraction of the primary direction of the ground plane. The RANSAC thresholds determine the distance from the emerging plane at which the points of the original point cloud should be considered inliers of the coarse ground.

At this stage, a significant number of points of the lower sections of the truss are located within the coarse ground and needs to be refined.

### 5.2 Fine Segmentation

The objective of the fine segmentation step is to enhance the previous segmentation. In this phase, the algorithm focuses on the coarse ground points.

In this phase of the algorithm a clustering process is applied. As the components of the target structures are composed of planar surfaces, region growing based on normals is employed to cluster planar segments in the cloud. The suitability of each cluster for inclusion in the truss is determined by examining its eigenvalues and eigenvectors. A prior knowledge of the bars that compose the structure (length and width) is used to evaluate whether a cluster belongs to the structure or not. The next subsections detail the steps of the algorithm.

### 5.2.1 Region Growing

The region growing algorithm is a clustering method based on the aggrupation of near points with similar features.

In order to divide the cloud into planar clusters, region growing with the normal vector as feature for each point is applied. The implementation of the region growing method employed in this study is based on the PCL library for C++. This method requires the normal vector of each point and its *Curvature* (Eq. (2)).

The growth of a region is initiated by the selection of an initial seed at the point of the lowest curvature (a point that resides on a plane) and subsequent evaluation of the normals of the nearest neighbours. If the specified threshold is met, the points in question are designated as new seeds, and the growth process continues.

Once the growing process exhausts all available seeds, a cluster is completed, and a new initial seed is established at the next point with the lowest curvature in the remaining cloud. This iterative process continues until the entire cloud has been evaluated.

*Normal Estimation and Curvature*

A precise estimation of the normal vector is of paramount importance for the efficacy of the clustering process. In this article, the estimation of normal vectors is based on Principal Component Analysis (PCA) applied to the covariance matrix of a point neighbourhood.

The covariance, denoted as $C$ is expressed in Eq. (1). Where $k$ is the total number of neighbours, $\boldsymbol{p}_i$ is the current neighbour point, $\overline{\boldsymbol{p}}$ is the centroid of the neighbourhood set, $\lambda_j$ is the eigenvalue, and $\vec{v}_j$ is the eigenvector for $j$.

The PCA over the covariance matrix returns the eigenvectors and eigenvalues of the clusters. The smallest eigenvector is considered the normal vector.

$$C = \frac{1}{k} \sum_{i=1}^{k} (\boldsymbol{p}_i - \overline{\boldsymbol{p}}) \cdot (\boldsymbol{p}_i - \overline{\boldsymbol{p}})^T$$
$$C \cdot \vec{v}_j = \lambda_j \cdot \vec{v}_j, \ j \in \{0, 1, 2\}$$

(1)

*Curvature* (Eq. (2)) is used to establish cluster boundaries. Since the eigenvalues ($\lambda_j$) are normalized between 0 and 1, the maximum curvature value is $\lambda_{max} = 1/3$, which occurs when $\lambda_0 = 1$, since $\lambda_0 \leq \lambda_1 \leq \lambda_2$ and $\lambda_{0-2} \in \{0, 1\}$. This explanation can be found in [59].

With regard to the curvature equation, it represents the sparsity of points along the normal vector. Points with low curvature indicate that their neighbouring points are situated on the same plane, whereas points with high curvature (maximun value 1/3) indicate that their neighbourhood is uniformly distributed in the three-dimensional space.

$$Curvature = \frac{\lambda_0}{\lambda_0 + \lambda_1 + \lambda_2}$$

(2)

### 5.2.2 Eigenvalues Evaluation

To determine if a cluster belongs to the structure or not, its eigenvalues and eigenvectors are evaluated.

The eigenvalues and eigenvectors of a cluster are obtained with a PCA analysis of the cluster (Eq. (1)). PCA results in three eigenvectors that follow the main directions of the spatial distribution of points, and its associated eigenvalues that represent the scalar factor of the vectors.

Three different ways of evaluating the eigenvalues have been established.

*Ratio*

This variant of the algorithm evaluates each cluster by the quotient of its two largest eigenvalues (Eq. (3)). This approach assumes that the elements of the structure have a slender and elongated geometry. The threshold for the ratio criterion can be established from the working principle of the algorithm and the dimensions of the bars in the structure. From the algorithm's principle, it could be deduced that the structure bar length should not exceed the RANSAC threshold employed in the coarse step, as only points with a distance to the ground plane lower than that value are fed to the fine segmentation. The bars' dimensions can be utilised to establish the maximum width at which planar clusters can be incorporated into the structure. The threshold for the ratio variant is then determined as the quotient between the width of the bars and the RANSAC threshold employed in the coarse segmentation, which is set to 0.3 (0.15 m/ 0.5 m) in this study.

$$ratio = \frac{\lambda_1}{\lambda_2} \tag{3}$$

*Magnitude*

The magnitude variant evaluates the length of the cluster along the largest eigenvector direction. This is done by estimating the eigenvectors of the cluster and applying a transformation to align the cluster basis with the eigenvectors directions. To set a cluster as part of the structure, its maximum dimension along the direction of the eigenvector $\lambda_2$, should be less than a given threshold. Like in the case of the ratio variant, the maximum length of the bars in the coarse segmented ground is the threshold employed for the ground plane estimation. For the purposes of this study, the value of this threshold has been set to 0.5 m.

*Hybrid*

The hybrid variant combines the two previous approaches in an attempt to exploit the strengths of both. For a cluster to be identified as a structure, it must meet both thresholds.

### 5.3 Density Filter

In the final step of the algorithm, and taking into account the proximity of the sensor to the structure, a density filter is applied to eliminate surrounding points. Points with low density are removed from the structure class. Although this filter could be applied at the beginning of the algorithm, preliminar tests showed that applying it after the fine segmentation process produced better results.

## 6 Experiments and Discussion

In order to ascertain the efficacy of the proposed methodologies, a series of experiments were conducted with the objective of evaluating their performance across a range of metrics. These experiments aimed to test the methods' capabilities in terms of mIoU, F1-score, efficiency, and reliability in different scenarios. Below, we outline the employed dataset, the key performance indicators and the experiments for both types of methods.

In this study, two computational setups were used. The first one, used for neural network training and testing consists of a computer with two Intel Xeon Gold 6334 CPUs with 128 gigabytes of RAM memory. The graphics processing unit (GPU) is an Nvidia GPU A30 Tensor Core with 24 GB of VRAM. The experiments were executed on the Ubuntu 22.04 operating system.

The second setup was used for the study of the analytical algorithm, which consists of a desktop computer with an Intel i7-10700K CPU, 32 GB of RAM, running under Ubuntu 20.04.

### 6.1 Datasets

Even though reticular structures are widely used in contemporary architectural designs, to the best of our knowledge, there is no widely established public database containing either images or metric data relevant to these environments. This absence of accessible information restricts the capacity to assess research efforts in this field.

Due to the lack of existing databases and the significant expense involved in constructing one from scratch, it is essential to employ an alternative approach for acquiring labelled 3D LiDAR datasets. In this study, as in others within the academic literature [31,32,60], we have opted for using a simulated environment to synthetically generate a sufficiently large and completely labelled 3D LiDAR dataset. An illustrative example of the simulated environment is provided in Fig. 4.

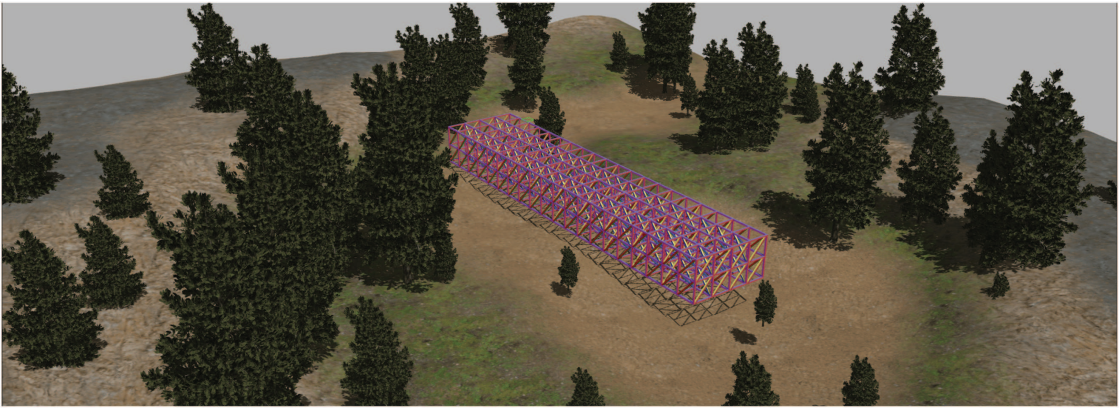

**Figure 4:** Example of the simulated environment for the dataset generation

The simulation platform employed is the Gazebo Simulator [61], an open-source application offering a comprehensive collection of robots and sensors, which is continually expanding due to the contributions of a broad community of developers. This simulator is integrated with the Robot Operating System (ROS), a widely recognised robotics software development framework that is extensively used in academic and research settings. Its integration with ROS, among other factors, has facilitated its widespread adoption and contributed to the size of its user community.

For this section, we drew on Sánchez et al. [60] research, in which automatic labelling of data from a LiDAR sensor is achieved using this simulator. The simulated sensor provides 4 fields of data per point, 3 for the cartesian coordinates of the point and one for the intensity related to that point. The default intensity value for each object in the environment is set to zero, but it can be manually adjusted. By assigning a distinct intensity value to each component of the environment intended for segmentation, this feature allows for the generation of a point cloud with fully labelled information.

In contrast to Sánchez et al. [60], where the pose of the sensors is manually selected for each reading, in our work, the process is entirely automated thanks to a plugin developed for this simulator, enabling the generation of a dataset in a fully automated manner. This plugin encompasses various functionalities, including the assignment of distinct intensity values to individual components within the environment, establishing random poses for designated elements within the environment, and saving point cloud information alongside its corresponding label in ".pcd" format. Each point cloud is generated independently, with its origin being the position of the sensor. To prevent issues during data capture, the plugin is manually restricted to generate only one scan per second.

Regarding the LiDAR sensor, Gazebo also permits the configuration of its operating parameters. The sensor used to generate the data has been parameterized in accordance with the configuration of the real 128-channel Ouster OS1 sensor. The parameters that model the behaviour of the simulated sensor are the resolution, the range of action, the vertical and horizontal fields of view and the noise applied to the data. In order to align the characteristics of the Ouster OS1 with those of the simulated sensor, the parameters outlined in Table 2 have been configured.

**Table 2:** Configuration parameters for the simulated sensor

| Parameter | Value |
| --- | --- |
| V. Resolution | 128 points |
| H. Resolution | 512 points |
| Min. Range | 0.050 m |
| Max. Range | 30.00 m |
| V. FOV. | 45° |
| H. FOV. | 360° |
| Noise | 0.008 m |

Two datasets have been built and used for this study. The first dataset contains the information used for training the neural network models, while the second contains the examples used to test the performance of the proposed methods. An overview of both datasets is detailed below.

### 6.1.1 Truss Generation Module

As part of this work, a program has been developed which is capable of generating generic structures to obtain a variety of environments to evaluate the proposed methods. Fig. 5 shows the structures generated with this module.

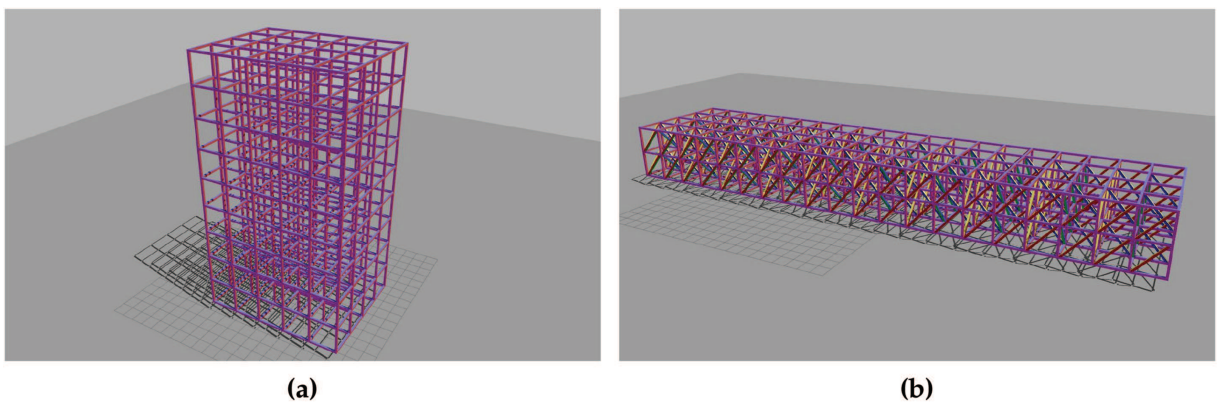

        **(a)**                                     **(b)**

**Figure 5:** Types of structures generated by the truss generation module: (**a**) Completely orthogonal structure; (**b**) Structure with crossed bars

These types of structures are modelled in an XML-based format called ".sdf". This format is native to the Gazebo software and is used to describe any element in the simulation environment. Such files allow the description of the element in question, encompassing both dynamic and kinematic properties, as well as visual aspects.

By using Gazebo's native format, the developed software allows the generation of structures in which each plannar face of a bar is described by a unique identifier. The LiDAR sensors implemented in Gazebo can read this information and assign the corresponding identifier to each point.

Thanks to this structure generation module, different examples with various dimensions and arrangements can be created. In its current state, this module allows the generation of two types of structures: those in which all elements are orthogonal to each other, and structures that also include inclined elements between the bars, simulating more intricate environments.

The program starts with a base model for the bars and replicates them, adjusting their dimensions to construct the desired structure. In the current state of the program, the base model is a cube, which is scaled along each of its axes directions based on the desired length and width of the bar. As the base model is a cube with a square cross-section, only two values are required to define the bar: the length and the width, which fully define its cross-section.

The configuration file allows the user to specify the number of bars in each of the three main directions ($X$, $Y$, $Z$). Furthermore, the appearance of the final structure (with or without texture) and the labelling process (label each bar or label each bar face) can be specified. A structure with a label per bar face results in a higher memory cost, but it enables a more fine-grained segmentation.

The base model determines whether the structure will exhibit a textured appearance, a label per bar, or a label on each planar face of the bar.

*6.1.2 Training*

In order to train the neural networks as broadly as possible, it was determined that a dataset should be generated containing parallelepipeds with dimensions and proportions that closely resemble those of the target structures.

Also, to ensure that the database is as comprehensive as possible, each parallelepiped is labelled differently on each of its planes. The label assigned to each plane is incrementally set for all planes belonging to the environment. This methodology enables the future retrieval of the ground truth regarding the segmentation of each individual plane, if necessary. If it is only necessary to distinguish between the parallelepipeds and the environment, it is sufficient to set all points with a label different from 0 as parallelepipeds. If no label is assigned to a model, a value of 0 is used by default.

The simulated environment is an outdoor terrain on which trees have been introduced to generate a more realistic scenario (Fig. 4). The database was generated using a variant of the developed plugin, which, in addition to modifying the position and orientation of the elements of the environment, allows the scale of the model to be modified along each of its axes individually. This enables the parallelepipeds to have different measures and orientations, covering a wide range of possible lengths and widths within a user defined scope. In this instance, the bars may vary in length and width between the specified limits. At the same time, the surrounding elements (trees) are also placed in random positions and their scale is modified.

In order to generate data, a process is employed that entails the maintenance of a constant sensor position while varying the orientation randomly. Next, an environment is created around the sensor, comprising parallelepipeds and trees exhibiting random pose and scale. Finally, the point cloud is stored under these circumstances, and a new random environment and sensor orientation are established.

For the training process, a dataset of approximately 10,000 point clouds has been generated. The dataset has been randomly partitioned into an 80% training set and a 20% validation set. The appearance of the segmentation ground truth of a training cloud can be appreciated in Fig. 6, in which the points that belong to the parallelepipeds are shown with green color and the rest of the points with black color.

Pytorch library has been employed for the training of all the models. PointNet, PointNet++, and MinkUNet34C adhere to the same training procedure, using the following parameters: loss function BinaryCrossEntropy, Adam optimiser, ReduceOnPlateau as scheduler with a reduction factor of 0.1, patience of 3 and threshold of 0.0001. No epoch limit has been set, however, the training is terminated when the model demonstrates no improvement in the loss function over 10 epochs. *Sigmoid* is applied as the activation function to the last layer of the model. As *Sigmoid* returns values between [0, 1], a threshold is estimated using the *Precision-Recall* (PR) curve or the *Receiver-Operating Characteristic* (ROC) curve to complete binary segmentation. This threshold is calculated during the validation phase and is subsequently employed for the testing phase.

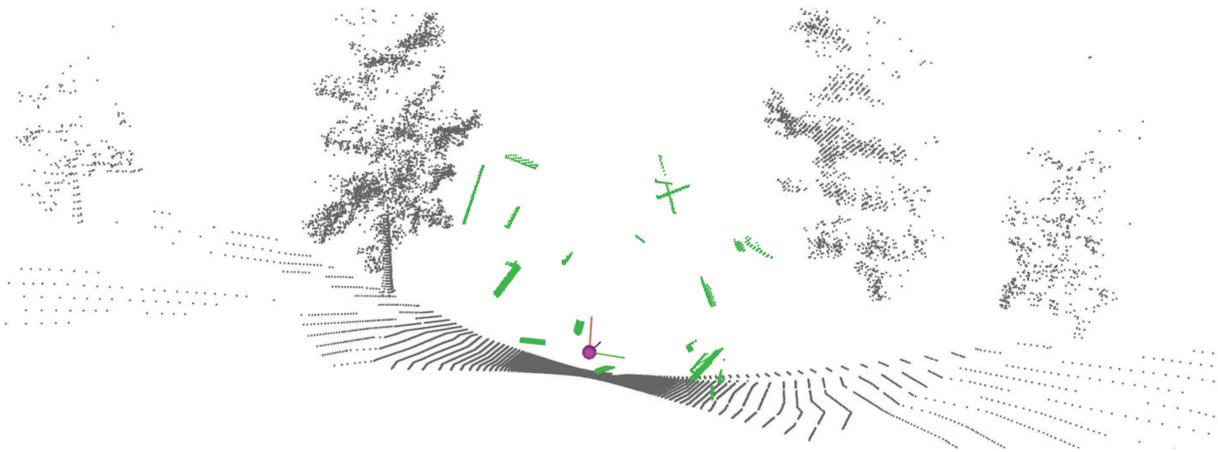

**Figure 6:** Example of training cloud. Sensor pose is represented by a purple sphere with the correspoinding axes

PointTransformerV3 has been trained using the same methodology as the original authors describe in [62] and does not need threshold estimation.

The MinkUNet34C and PointTransformerV3 models have been trained with a discretization size of 0.05 m. The coordinates of the input vector are left unnormalized for all models except PointNet++.

### 6.1.3 Test

Realistic reticular structure models are used for the generation of the test dataset. Two structures were created with the *Truss Generation Module* (Section 6.1.1): one with orthogonal elements (Fig. 5a) and a more complex structure featuring crossed elements (Fig. 5b). The dimensions of the crossed structure are 40 m × 8 m × 4 m (length, width and height, respectively). The orthogonal truss measures 10 m × 8 m × 18 m. In both structures, the horizontal and vertical bars are 2.0 m in length and 0.15 m in width.

For the automatic generation of single scans, the structure and the environment are static, and the developed plugin moves the sensor pose randomly within the range defined by the dimensions of the structure. For each structure (orthogonal and crossed) 1000 point clouds were generated.

### 6.2 Metrics

To measure the success of the experiments, we defined key performance indicators for each test. The selected indicators the F1-score, mIoU and computation time. In this section, these metrics are outlined.

The metrics employed are derived from the calculation of the confusion matrix, which represents the ground truth in relation to the estimated segmentation. This matrix contains the values: True Positives (TP), False Positives (FP), True Negatives (TN) and False Negatives (FN).

*F1-score*

From the values in the confusion matrix, it is possible to estimate the *Precision* (Eq. (4)) and *Recall* (Eq. (5)), which respectively represent the degree of confidence of the model and the percentage of data that the model is able to correctly segment. These metrics can be combined simultaneously to form the *F1 score* (Eq. (6)), which provides a more accurate representation of the model's overall performance.

$$Precision = \frac{TP}{TP + FP} \tag{4}$$

$$Recall = \frac{TP}{TP + FN} \tag{5}$$

$$F1 = 2 \times \frac{Precision \times Recall}{Precision + Recall} \tag{6}$$

*Mean Intersection over Union (mIoU)*

This metric is widely used for the evaluation of segmentation tasks. It evaluates the overlap between the prediction and the ground truth. It measures how well the model can separate both classes, truss and ground.

$$mIoU = \frac{TP}{TP + FP + FN} \tag{7}$$

### 6.3 Results

This section shows the results of a bunch of experiments carried out for each segmentation method. Moreover, the sensitivity of the methods is studied considering different types of input features (for the deep learning methods) or the different ways of evaluating the eigenvalues for the analytical method.

The results of the analytical method are evaluated considering both orthogonal and crossed datasets. In the Table 3, three eigenvalues evaluation modes are considered: **R** (Ratio), **M** (Magnitude), **H** (Hybrid). Additionally, the performance of the algorithm is studied if only one of the two steps is run: **WF** (Without Fine Segmentation), and **WC** (Without Coarse Segmentation). WC followed by the sufix, R, M or H, means that only the fine step is applied for the whole cloud, with the corresponding variant of the fine step ratio, magnitude or hybrid. The values representing the F1-score and mIoU are calculated as the mean value of all of the corresponding dataset point clouds.

**Table 3:** Performance of different variants of the analytical method considering either orthogonal or crossed structure datasets. The best results are highlighted in bold

| Mode | Dataset | F1-score | mIoU |
|------|---------|----------|------|
| R | ortho | 84.62% | 73.34% |
| M | ortho | 89.77% | 81.43% |

(Continued)

**Table 3 (continued)**

| Mode | Dataset | F1-score | mIoU |
|---|---|---|---|
| H | ortho | **89.78%** | **81.45%** |
| WF | ortho | 89.64% | 81.22% |
| WC_R | ortho | 51.56% | 34.73% |
| WC_M | ortho | 54.48% | 37.43% |
| WC_H | ortho | 51.91% | 35.06% |
| R | crossed | 79.59% | 66.10% |
| M | crossed | **93.16%** | **87,19%** |
| H | crossed | 93.13% | 87.14% |
| WF | crossed | 93.11% | 87.10% |
| WC_R | crossed | 44.10% | 28.29% |
| WC_M | crossed | 52.44% | 35.54% |
| WC_H | crossed | 40.64% | 25.50% |

The voxel leaf size, the RANSAC threshold, and the density filter parameters were empirically determined and consistent across all experiments: 0.1 m for the voxel leaf size, 0.5 m for the RANSAC threshold and a search radius of 0.25 m with a threshold of 10 points for the density filter.

The results show comparable outcomes for the three variants, with the magnitude and hybrid exhibiting superior performance for both datasets. The distinction between magnitude and hybrid variants is not significant, due to two factors. Firstly, the limited data on which the fine segmentation is applied. Secondly, the greater tendency of the hybrid variant to be more restrictive. This results in fewer positives being detected. The choice of variant depends on the distribution of points in the dataset. In this case, the orthogonal dataset has a higher number of negative points during fine segmentation than the crossed dataset.

As fine segmentation is designed to identify small regions of the structure with few points, the difference in results between applying and not applying fine segmentation is minimal, typically within the range of 0.05%–0.2%.

Additionally, the configurations that solely use fine segmentation across the entire cloud, regardless of the eigenvalues evaluation mode, yield diminished outcomes. This is attributed to the rigorous nature of the thresholds and the calculation of the eigenvector for each set.

The limitations of the ad hoc algorithm can be traced back to two primary sources. Firstly, the clustering of planar pathways requires an accurate normal estimation to ensure effective segmentation through region growing techniques. This process requires precise parameter adjustment to ensure optimal outcomes.

The second limitation comes from the occlusions and partial views of the bars as they are perceived from the sensors. The consequence of the aforementioned scans is the potential for miscalculation of the two largest eigenvectors. Occlusions may result in an eigenvector that is not aligned with the bar's primary direction. This could lead to an inaccurate estimation of the bar's dimensions, with misclassification of the points in the structure. This limitation could be observed in the variants of the algorithm that only use fine segmentation, achieving poorer results due to misalignment of the eigenvectors with the bars.

The disparities between the two datasets, i.e., the orthogonal and the crossed, are attributable to the density of the structure and the inherent characteristics of the employed algorithm. As previously mentioned in Section 6.1.3, the crossed structure exhibits a lower height, higher density and more occlusion of points at greater distances. This configuration is advantageous to the algorithm in terms of the density filter, as the points of the structure are closer to the sensor, resulting in a more compact and dense point cloud. Conversely, the orthogonal structure with greater height, fewer bars and reduced occlusions, results in a sparser and less dense configuration which increases the likelihood of misclassifications when using the density filter.

The outcomes of the analytical method are encouraging, with both the magnitude and hybrid variants demonstrating optimal performance. The findings suggest that the algorithm performs more effectively in dense structures with a higher number of bars. Nevertheless, the algorithm relies on point clouds containing a substantial number of points on a surface to accurately estimate the direction along which the structure emerges. This surface does not necessarily have to be the ground, it could also be the roof of a bridge or a wall. Nonetheless, the algorithm depends on the structures emerging from some surface to ensure accurate identification.

A summary of the results obtained using deep learning methods is presented in Fig. 7. This figure shows the performance of the optimal model with regard to each deep learning model and type of input feature. In this experiment different combinations of input features are considered: **V** represents the 3D coordinate vector ($x,y,z$), **N** represents the normal vector ($nx,ny,nz$), and **C** represents the curvature. Two methods for estimating the threshold are employed for each dataset, the *Receiver-Operating Characteristic* (ROC) curve (righ-hand), and the *Precision-Recall* (PR) curve (left-hand).

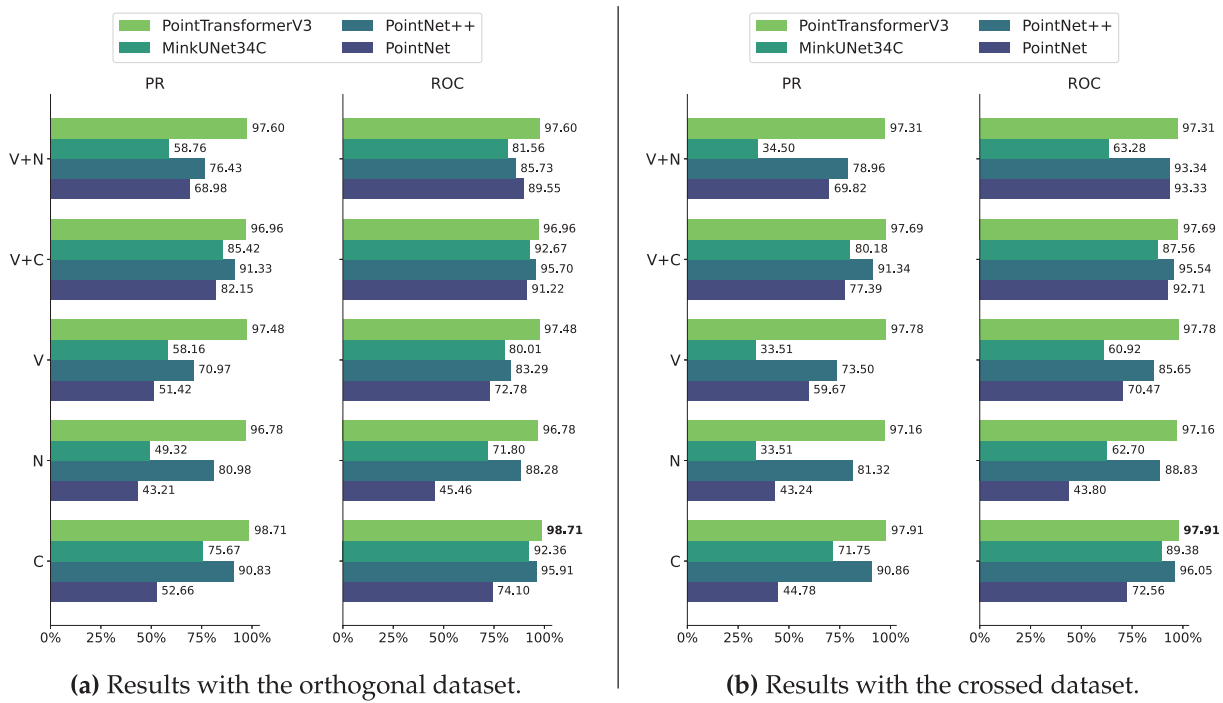

**(a)** Results with the orthogonal dataset.

**(b)** Results with the crossed dataset.

**Figure 7:** Mean intersection over union (mIoU) for each deep learning method. (**a**) shows the results with the orthogonal dataset, and (**b**) with the crossed dataset. Best results are highlighted in bold

The ROC curve plots the *True Positive Rate* (TPR) against the *False Positive Rate* (FPR), illustrating the classifier's performance in differentiating between classes. The optimal threshold is chosen at the point where the geometric mean, defined as **gmean** = $\sqrt{\mathbf{TPR} \cdot (1 - \mathbf{FPR})}$, reaches its maximum value. Here, **TPR**, **FPR** $\in \mathcal{R}^x$, and $x$ is the number of thresholds tested.

The PR curve plots the Precision against Recall at different thresholds. For each pair of Precision and Recall, the F1-score is calculated. The optimal threshold is selected to maximize the F1-score.

As discussed in Section 6.1.2, PointTransformerV3 is not influenced by the threshold, its metrics are duplicated for PR and ROC threshold estimation to improve the figure's symmetry.

Fig. 7 reveals that the models that use ROC curve threshold estimation perform better than PR curve across all cases, positioning the ROC curve as a more robust and efficient estimator to achieve best mIoU results in planar binary segmentation.

Leaving aside the threshold estimation and focusing on the input features, the Fig. 8 provides some key findings. First, the curvature (**C**) or the combination of it with other features (**V+C**) positions it as the best feature for characterizing planar environments and discretizing planar features with neural networks. PointNet++, MinkUNet34C and PointTransformerV3 achieve their best mIoU with this input feature.

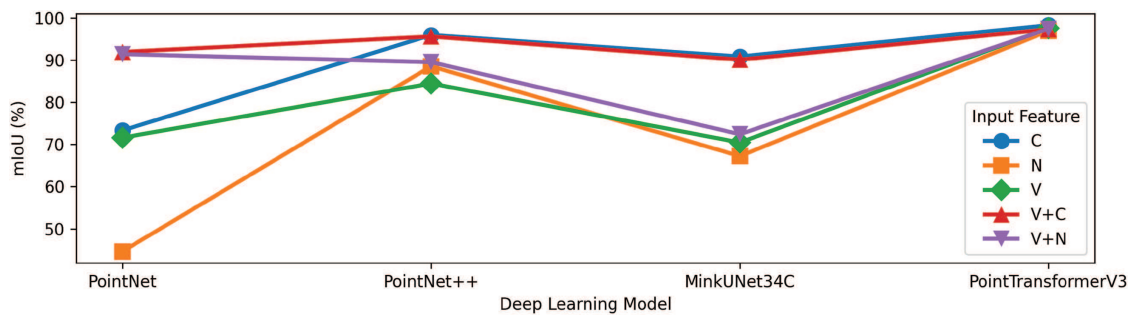

**Figure 8:** Deep Learning model against mIoU for each input feature. The denomination of the input features is the same as in the previous figure

For PointNet model, the absence of spatial coordinates in the input reduces considerably the performance of the model, for example using just the curvature achieves 73% of mIoU and this metric grows to 91% in combination with spatial coordinates. The low performance of PointNet without spatial coordinates could be also seen with the normal vector as input (**N**) with a minimal performance of 44% mIoU.

Moving on to the models, PointNet has proved to significantly improve the results by combining normal vector with spatial coordinates (V+N), highlighting its proficiency in exploiting spatial and geometrical features. In comparison to other models, PointNet++ demonstrates reduced sensitivity to input features, achieving a comparable mIoU with different input features. The MinkUNet34C model demonstrated suboptimal performance in this study, a factor that could be affected by the voxel size used during the training phase (0.05 m), which may have been too small for the model to capture the details of the structure.

Finally, PointTransformerV3 emerges as the top deep learning model for binary segmentation in structures achieving an mIoU of 98.31% as mean of both datasets, followed by PointNet++ with 95.98% in the same situation. Furthermore, an analysis of this model reveals that it demonstrates a reduced sensitivity to the input features when compared to alternative models. The findings suggest that PointTransformerV3 is the optimal model for binary segmentation of planar structures, as evidenced by the absence of threshold estimation, independence of input features and the high mIoU results.

Fig. 9 visually shows the results obtained with a sample point cloud captured from **(a)** the orthogonal dataset and **(b)** the crossed dataset. The ground truth of the segmentation is presented in the first row, the second row exhibits the results of the analytical method, where the hybrid solution is employed for the orthogonal dataset (a) and the magnitude solution for the crossed dataset (b). The third row represent the predictions from PointTransformerV3, using curvature as input feature. In each figure, the sensor pose is represented by a purple sphere and its corresponding axes.

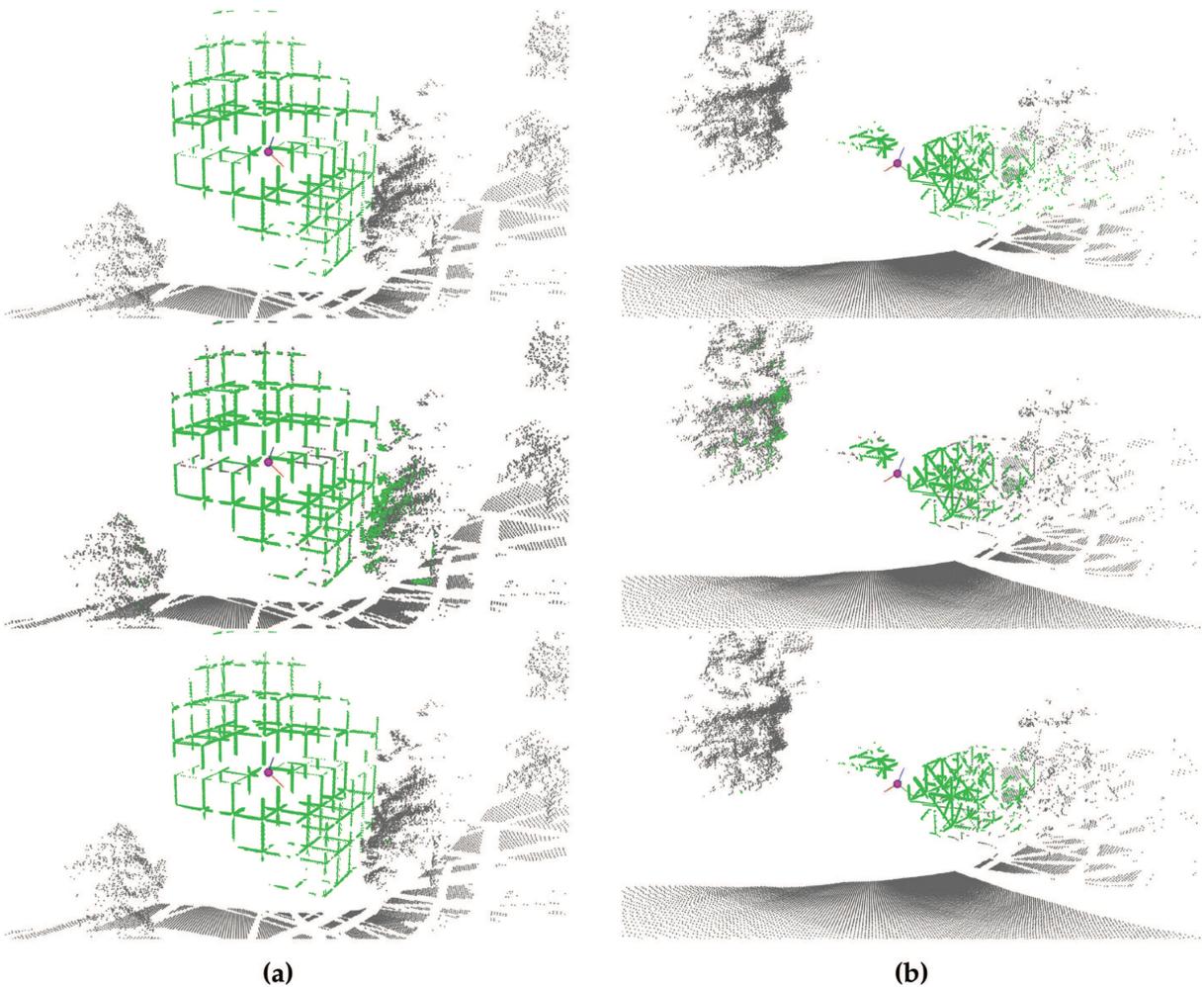

**(a)**                                                    **(b)**

**Figure 9:** Performance comparison between the hybrid analytical method and PointTransformerV3, across different datasets. (**a**) Orthogonal dataset; (**b**) Crossed dataset

The results demonstrate the remarkable capacity of PointTransformerV3 for data segmentation, with a representation that closely aligns with the ground truth. Conversely, the analytical method yields lower segmentation results, with satisfactory segmentation of the structure but exhibiting challenges in segmenting regions proximate to the sensor that do not belong to the structure (nearby trees).

In addition, the performance of PointTransformerV3 is analysed considering several models of realistic structures formed by elements completely unknown to the neural network (Fig. 10 first column).

The evaluation of PointTransformerV3 with other different structures demonstrates highly satisfactory results. It exhibits robust performance on structures comprising diverse bar types and is even capable of differentiating the roof from the bars. The process of segmentation begins to present difficulties in areas situated at a notable distance from the sensor. Visual results are shown in Fig. 10.

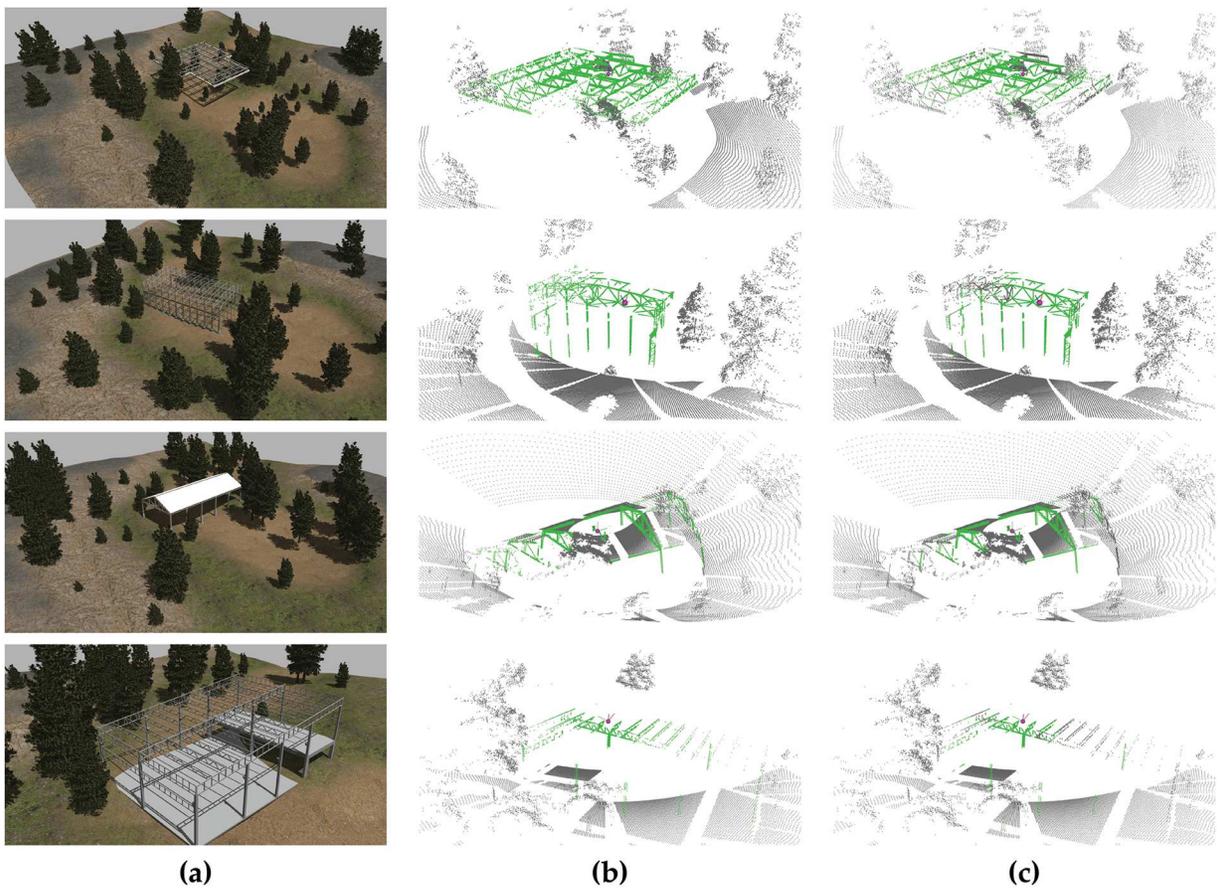

        **(a)**                             **(b)**                           **(c)**

**Figure 10:** Inferences of PointTransformerV3 over complex additional structures. Column (**a**) represents the environment; (**b**) shows the ground truth segmentation; (**c**) shows the inference of the best PointTransformerV3 model

The encouraging results of PointTransformerV3 can be attributed to its key distinction from other models: the utilisation of transformers in place of conventional convolutions. These blocks incorporate internal attention mechanisms, which possess a remarkable capability to discern intricate correlations between the input data. This positions them as a leading choice for semantic segmentation within the current state of the art.

Notwithstanding the promising results obtained with neural networks, limitations within the deep learning paradigm persist. As demonstrated in Fig. 11, two analogous point clouds yield divergent outcomes when processed by the PointTransformerV3 neural network model. This discrepancy is difficult to justify, given the comparable sensor location, point configuration, and density in the regions with the classification errors. While it is possible to theorise about the causes of this behaviour, the solution is to conduct numerous training sessions, varying the model architecture or input features through a trial-and-error approach. This behaviour is a common occurrence in the deep learning paradigm, where the model acts as an opaque system, yielding a limitation in the control and results interpretation.

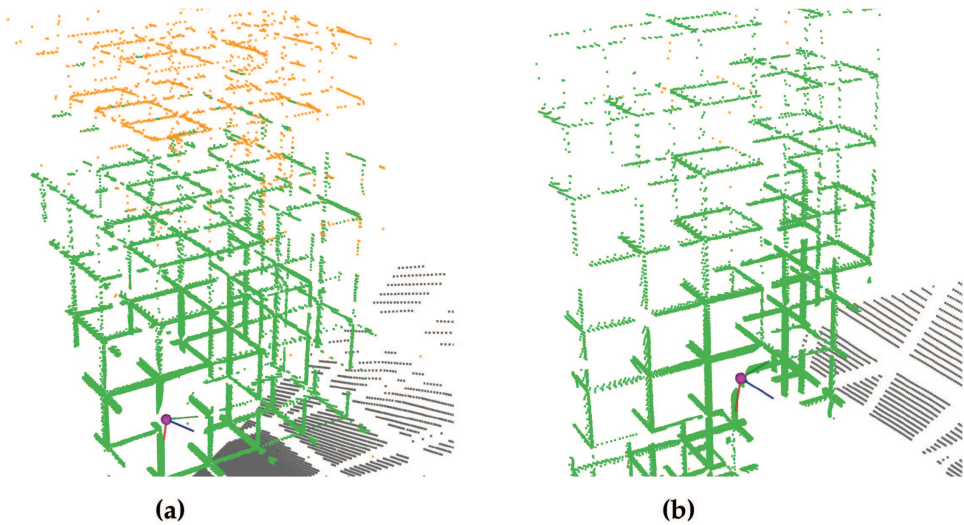

**Figure 11:** Inconsistency of neural network models against similar point clouds. Examples of inferences of PointTransformerV3; (**a**) Point cloud with lower mIoU (93.52%); (**b**) Point cloud with higher mIoU (99.86%). Orange points represents misclassifications

Finally, to assess the efficiency of each method, the execution time and memory consumption are evaluated. The metrics for each model are computed considering the same conditions, with a batch size of 4 for all the models and the memory used after the first epoch. To calculate the metrics the best model of each method for the orthogonal dataset is employed.

Memory value corresponds to reserved memory in the CUDA device given from Pytorch. For the analytical method, only inference metrics are established. The results of this analysis are shown in Table 4. The analytical method is the most efficient in terms of memory as expected.

**Table 4:** Latency and required memory of the segmentation methods. PointNet (**PN**), PointNet++ (**PN++**), MinkUNet34C (**MKN34C**), PointTransformerV3 (**PTV3**), Analytical method with the Hybrid variant (**H**). Inference results for the orthogonal dataset. The best results for each metric are highlighted in bold

| Methods | Training | | Inference | |
|---------|----------|--------|-----------|--------|
| | **Latency** | **Memory** | **Latency** | **Memory** |
| PN [40] | **15 ms** | 3.68 Gb | **117 ms** | 0.41 Gb |
| PN++ [37] | 82 ms | 4.34 Gb | 345 ms | 1.11 Gb |
| MKN34C [52] | 41 ms | **1.56 Gb** | 260 ms | 0.36 Gb |
| PTV3 [53] | 105 ms | 16.18 Gb | 599 ms | 0.72 Gb |
| H | – ms | – Gb | 212 ms | **0.09 Gb** |

As demonstrated in Table 4, PointTransformerV3 is not the most efficient model in terms of memory and latency. This finding indicates that despite the enhanced efficiency of transformers blocks, simple convolutions prove to be more efficient in terms of latency and memory. The deployment of models such as PointTransformerV3 may be restricted when operated on mobile devices, including climbing robots, due to its requirement for greater memory and computing capabilities compared to alternative models.

Conversely, the analytical method has been shown to produce satisfactory results with fewer resources, making it more suitable for small devices.

With regard to the application of these methods to real-world data, a decrease in performance is to be expected due to the limitations arising from the simulation. However, it is considered to be a sufficiently realistic simulation, since it includes data noise, LiDAR occlusions and realistic environment models, so that the main limitation will depend on the environment in which it is applied. Real-world environments similar to those used in this work should not present issues when applying any of the proposed methods. It is hypothesised that the performance of more generalist methods, such as PointTransformerV3, will exceed that of more specialized approaches, including the analytical approach, as demonstrated in Fig. 10. Nevertheless, the analytical method may offer enhanced results due to its flexibility in adjusting parameters and adapting to specific environments. The selection of the method to be used will depend on the characteristics of the environment and the resources available on the robot.

## 7 Conclusions

This work introduces several methods for the segmentation of lattice structures. Due to the limited availability of databases for such structures, simulated environments have been employed to obtain data. A custom plugin for Gazebo has been developed to automatically generate labelled 3D LiDAR datasets. Furthermore, a software module has been created to automatically generate structures based on parameters, thus enabling the assessment of the methods across diverse environments with minimal development cost.

Two distinct approaches have been employed to address the issue of segmentation. The initial approach employed analytical algorithms while the second one used deep learning techniques. In the evaluation of the analytical paradigm, the hybrid method has presented optimal results for mIoU, with a mean value of 84.32%. The findings indicate that this method is the most efficient in terms of memory, and the second most efficient in terms of execution time. However, its performance is constrained by the precision of the eigenvectors calculation, whose misalignment with the real principal direction of the bars may lead to erroneous bar classification. Additionally, the method is hindered by the challenge of accurately classifying points that are distant from the sensor, due to the low density of these points. Notwithstanding these limitations, the method's efficiency, parameters accessibility, low resource requirements, such as large amounts of data or dedicated hardware, and its overall performance make it a compelling option for small robots with limited capacities or for situations where a fast adjustment of the operation to a specific environment is necessary.

With regard to the application of deep learning methods, this study provides a compelling demonstration of the viability of employing generalised training, initiated exclusively from parallelepipeds, to achieve a satisfactory segmentation of intricate elements such as reticular structures. Furthermore, it is evident that these methodologies consistently yield superior performance in comparison to analytical approaches.

Another noteworthy aspect pertains to the estimation of the decision threshold in PointNet, PointNet++ and MinkUNet34C models, wherein the ROC curve exhibits an average increase of approximately 12% in mIoU in comparison to the estimation of the threshold with PR curves. These models also demonstrate a certain degree of dependence on the input features, with the utilisation of curvature, in conjunction with other features, proving to be the optimal approach for binary segmentation of reticular structures, resulting in an enhancement of 10–20% in mIoU compared to alternative features.

In the realm of deep learning methodologies, PointTransformerV3 has emerged as the preeminent model for the segmentation of truss structures. It has attained a maximum mIoU of 98.31%, which surpasses the second-best PointNet++ model by 2.33%. The superiority of PointTransformerV3 can be attributed to several factors, including the incorporation of transformers, the absence of threshold estimation, and its independence from the input features. This results in superior performance in terms of mIoU and better generalisation to new environments and complex structures. Nevertheless, its excellent performance in segmentation is constrained by its substantial demand for resources, both in terms of memory and execution time, being the second most memory intensive model.

Another key limitation of these methods is the lack of consistency between results with very similar inputs. The opacity of deep learning models makes it difficult to justify and interpret the results obtained. In such cases, modifying the model behaviour may require multiple training sessions with adjustments to the architecture and input parameters. This process can be costly and time-consuming. This stands in contrast to analytical algorithms, where the behaviour of the algorithm can be interpreted and addressed through logical and reasoned solutions devised by researchers.

The efficiency of the methods is also evaluated, demonstrating that the analytical method is considerably more efficient than deep learning models, in terms of memory and latency. In certain circumstances and with respect to specific tasks, it may be advantageous, or even essential, to employ more efficient methods that can be executed on robotic systems with constrained resources.

The choice of the most appropriate methodology is dependent upon the particular circumstances under consideration, with factors such as efficiency, the accuracy of segmentation, and the availability of resources on the robot.

In future studies, the verification of the methods with real datasets could be a valuable contribution to the field. However, this may present challenges when the environment differs from the ones used in simulation, e.g., indoor environments. In such cases, new datasets and training procedures should be conducted for neural network models and parameter adjustments of the ad hoc algorithm should be done. Future improvements could also include the truss generation module by integrating alternative base models, such as cylinders or more complex bar representations, to facilitate the creation of more detailed and accurate structural models. Furthermore, the extension of the proposed methods to operate with more cost-effective sensors, such as RGB-D cameras, could be explored with a view to improving accessibility and reducing implementation costs. Another potential direction for research involves the utilisation of a per-face labelled dataset to develop methods for instance segmentation of bar surfaces. This would address not only the segmentation between the structure and background, but also the detection of individual attachable surfaces. The identification of individual planes could be exploited to construct lightweight parametric representations, enabling highly scalable maps in large-scale environments.

**Acknowledgement:** The authors gratefully acknowledge the support of the Engineering Research Institute of Elche (I3E) and the University Miguel Hernández of Elche (UMH).

**Funding Statement:** This work has been funded by the spanish Ministry of Science, Innovation and Universities as part of the project PID2020-116418RB-I00 funded by MCIN/AEI/10.13039/501100011033.

**Author Contributions:** The authors confirm contribution to the paper as follows: All authors contributed equally to the conceptualization and design of the study; methodology: Luis Payá Castelló, Óscar Reinoso García; resources: Adrián Peidró Vidal, Marc Fabregat-Jaén; analysis and results interpretation: Francisco J. Soler Mora; data collection: Francisco J. Soler Mora; writing—original draft preparation: Francisco J. Soler Mora; writing—review and editing: Francisco J. Soler Mora, Luis Payá Castelló, Óscar Reinoso García; supervision: Luis Payá Castelló, Óscar Reinoso García; funding acquisition: Luis Payá Castelló, Óscar Reinoso García. All authors reviewed the results and approved the final version of the manuscript.

**Availability of Data and Materials:** The data that support the findings of this study are openly available in the GitHub repository rs_seg_methods at https://github.com/Urwik/rs_seg_methods.git (accessed on 20 May 2025).

**Ethics Approval:** Not applicable.

**Conflicts of Interest:** The authors declare no conflicts of interest to report regarding the present study.